\documentclass[sigconf]{acmart}

\usepackage{booktabs}
\usepackage{amsmath}
\usepackage[linesnumbered,ruled]{algorithm2e}
\usepackage{multicol}
\usepackage{enumitem}
\setlist[enumerate]{itemsep=0mm}

\newcommand{\minisection}[1]{{\vspace{3pt}\noindent\textbf{#1.}}}

\setlist{leftmargin=4.5mm}
\renewcommand\footnotetextcopyrightpermission[1]{}
\begin{document}
\title{Real-Time Bidding with Multi-Agent Reinforcement Learning\\in Display Advertising}

 \author{Junqi  Jin$^{\S}$, Chengru Song $^{\S}$, Han Li$^{\S}$, Kun  Gai$^{\S}$, Jun Wang$^{\dag}$,  Weinan Zhang$^{\ddag}$
    \\
{\large $^{\S}$Alibaba Group, $^{\dag}$University College London, $^{\ddag}$Shanghai JiaoTong University}\\
{\normalsize\{junqi.jjq, chengru.scr, lihan.lh, jingshi.gk\}@alibaba-inc.com, j.wang@cs.ucl.ac.uk, wnzhang@sjtu.edu.cn}\\
}

\renewcommand{\shortauthors}{}

\begin{abstract}
Real-time advertising allows advertisers to bid for each impression for a visiting user.
To optimize specific goals such as maximizing revenue and return on investment (ROI) led by ad placements, advertisers not only need to estimate the relevance between the ads and user's interests, but most importantly require a strategic response with respect to other advertisers bidding in the market.
In this paper, we formulate bidding optimization with multi-agent reinforcement learning. To deal with a large number of advertisers, we propose a clustering method and assign each cluster with a strategic bidding agent.  A practical Distributed Coordinated Multi-Agent Bidding (DCMAB) has been proposed and implemented to balance the tradeoff between the competition and cooperation among advertisers. The empirical study on our industry-scaled real-world data has demonstrated the effectiveness of our methods. Our results show cluster-based bidding would largely outperform single-agent and bandit approaches, and the coordinated bidding achieves better overall objectives than purely self-interested bidding agents.
\end{abstract}



\maketitle

\textbf{Reference Format:}\\
Junqi Jin, Chengru Song, Han Li, Kun Gai, Jun Wang, Weinan Zhang. 2018. Real-Time Bidding with Multi-Agent Reinforcement Learning in Display Advertising. In 2018 ACM International Conference on Information and Knowledge Management (CIKM '18), October, 2018, Torino, Italy.

\section{Introduction}\label{section:Introduction}

Online advertising \cite{evans2009online,goldfarb2011online} is a marketing paradigm utilizing the Internet to target audience and drive conversions. Real-time bidding (RTB) \cite{wang2017display} allows advertisers to bid for every individual impression in realtime when being generated. A typical RTB ad exchange employs the second price sealed-bid auction \cite{,yuan2014empirical}, and in theory (under strong assumptions) the second price auction would encourage truthful bidding.
In practice, however, the optimal or equilibrium bids are largely unknown, depending on various factors, including the availability of market bid prices, the existence of budget constraints, performance objectives, (ir)rationality of opponent bidders. As such, how to \emph{strategically} optimize bidding becomes a central question in RTB advertising \cite{yuan2013real}.

The research on optimal bidding strategies so far has been focused largely on statistical solutions, making a strong assumption that the market data is \emph{stationary} (i.e. their probability distribution does not change over time in response to the current bidder's behaviors) \cite{cai2017real,wang2017ladder,zhu2017optimized,perlich2012bid,zhang2014optimal}. Specially, \citet{zhang2014optimal} shows that budget-constrained optimal bidding can be achieved under the condition that the environment (along with other ad bidders) is stationary.  \citet{zhu2017optimized} proposes a two-stage bandit modeling where each bidding decision is independent over time. \citet{cai2017real} and \citet{wang2017ladder} leverage reinforcement learning to model the bid optimization as a sequential decision procedure. Nonetheless, in ad auctions, ad campaign bidders not only interact with the auction environment but, most critically, with each other. The changes in the strategy of one bidder would affect the strategies of other bidders and vice versa \cite{tan1993multi}. In addition, existing computational bidding methods \cite{perlich2012bid,zhang2014optimal} are mainly concerned with micro-level optimization of one party (a specific advertiser or merchant)'s benefit. But given the competition in the RTB auction, optimizing one party's benefit may ignore and hurt other parties' benefits. From the ad system's viewpoint, the micro-level optimization may not fully utilize the dynamics of the ad ecosystem in order to achieve better social optimality \cite{zhu2017optimized,wang2017ladder}.

In this paper, we address the above issue by taking a game-theoretical approach \cite{nisan2007algorithmic}. RTB is solved by multi-agent reinforcement learning (MARL) \cite{hu1998multiagent},  where bidding agents interactions are modeled. A significant advantage over the previous methods \cite{cai2017real,wang2017ladder,zhu2017optimized,perlich2012bid,zhang2014optimal} is that our proposed MARL bidding strategy is \emph{rational} as each bidding agent is motivated by maximizing their own payoff; it is also \emph{strategic} as each bidding agent will also provide a best response to the strategic change of other bidders to eventually reach to an equilibrium stage.

Our study is large-scale and developed in the context of a realistic industry setting, Taobao (\url{taobao.com}), the largest e-commerce platform in China. Taobao serves over four hundred million active users. The RTB exchange itself serves more than one hundred millions active audiences every single day. To our best knowledge, this is the first study of employing MARL for such large scale online advertising case, evaluated over real data. Previous studies on MARL are mostly in theoretical nature, and the majority experiments are done by simulated games \cite{lowe2017multi,hu1998multiagent}. Our RTB can be considered one of the earliest realistic applications of MARL.

Modeling large scale bidding by MARL is, however, difficult. In Taobao e-commerce platform, there are a large number of consumers and merchants. Modeling each merchant as a strategic agent is computationally infeasible. To tackle this issue, we propose that bidding agents operate in the clustering level. We cluster consumers into several groups, each of which is considered as a "super-consumer", and also cluster merchants into groups, each of which is represented by a common bidding agent.
The multi-agent formulation is thus based on the interactions between super-consumers and cluster-level bidding agents, as well as the interactions among bidding agents. A technical challenge is the convergence of MARL as all the cluster bidding agents explore the auction system simultaneously, which makes the auction environment non-stationary and noisy for each agent to learn a stable policy. Inspired by multi-agent deep deterministic policy gradient (MADDPG) techniques \cite{lowe2017multi}, we propose Distributed Coordinated Multi-Agent Bidding (referred as DCMAB) method to stabilize the convergence by feeding all agents' bidding actions to the Q function. During learning, each bidding agent's Q function evaluates future value according to all agents' actions rather than only itself's action.

Our solution is fully distributed, and has been integrated with Taobao's distributed-worker system, which has high-concurrency and asynchronous requests from our consumers. Experiments are conducted on real world industrial data. The results demonstrate our DCMAB's advantage over several strong baselines including a deployed baselines in our system. We also find that when bidding agents act from only self-interested motivations, the equilibrium that converged to  may not necessarily represent a socially optimal solution \cite{leibo2017multi,wang2003reinforcement}. We thus develop a fully coordinated bidding model that learns the strategy by specifying a common objective function as a whole. The empirical study shows our DCMAB's ability of making merchants coordinated to reach a higher cooperative goal.
\section{Related Work}\label{section:Related_work}

\textbf{Bid Optimization in RTB.} Bidding optimization is one of the most concerned problems in RTB, which aims to set right bidding price for each auctioned impression to maximize key performance indicator (KPI) such as click or profit \cite{wang2017display}.
Perlich et al. \cite{perlich2012bid} first introduced a linear bidding strategy based on impression evaluation, which has been widely used in real-word applications. Zhang et al. \cite{zhang2014optimal} went beyond linear formulation. They found the non-linear relationship between optimal bid and impression evaluation. These methods regard bidding optimization as a static problem, thus fail to deal with dynamic situations and rationality of bidding agents.

More intelligent bidding strategies optimize KPI under certain constraints and make real-time adaption, most of which are met with reinforcement learning. Cai et al. \cite{cai2017real} used a Markov Decision Process (MDP) framework to learn sequentially allocating budget along impressions. Du et al. \cite{du2017improving} tackled budget constraint by Constrained MDP. Wang et al. \cite{wang2017ladder} utilized deep reinforcement learning, specifically DQN, to optimize the bidding strategy. They set high-level semantic information as state, and consider no budget constraint. These tasks share a common setting, i.e., bid optimization serves for one single advertiser, with its competitors as part of the environment, which significantly differs from our settings.

Another popular method for budget allocation is the pacing algorithm \cite{lee2013real,jxu2015smartpacing} which smooths budget spending across time according to traffic intensity fluctuation. Compared with our method, pacing can be considered as a single agent optimization method which does not explicitly model the influence from other agents' actions in the auction environment. In addition, pacing cannot coordinate agents to cooperate for a better equilibrium.

Like many other ad exchanges, in Taobao advertising system, we treat advertisers equally. Meanwhile, we need to balance the interests among consumers, advertisers and the platform. Thus, we are motivated to construct a framework that simultaneously takes different interests into consideration. Advertisers compete for high quality impressions, while they should cooperate in the sense of providing better user experience. In our work, we adopt multi-agent reinforcement learning to achieve this goal.

\minisection{Multi-agent Reinforcement Learning} In multi-agent literature, how to design mechanisms and algorithms to make agents well cooperate is the focus. Tan \cite{tan1993multi} compared cooperation with independent Q-learning, drawing the conclusion that additional information from other agents, if used properly, is beneficial for a collective reward. Many studies afterwards focused on how to effectively coordinate agents to achieve the common goal, either by means of sharing parameters \cite{gupta2017cooperative} or learning communication protocol \cite{foerster2016learning,mordatch2017emergence}. Some of these studies \cite{gupta2017cooperative,foerster2016learning} adopted the framework of centralized training with decentralized execution, allowing for involving extra information to ease training. Lowe et al. \cite{lowe2017multi} studied further in this direction and proposed MADDPG (Multi-agent DDPG), in which the centralized critic is augmented with policies of other agents. However, MADDPG was applied in a toy simulation environment where the states update and transition tuple saving can be performed frequently.

The most serious challenge in our task is that there are a huge number of advertisers in Taobao, which exceeds the processing capacity of almost all current multi-agent reinforcement learning methods. If we model each advertiser as an individual agent, the reward would be sparse for most agents. Besides, our bidding system is implemented on distributed workers which process requests in parallel and asynchronously. Considering all these factors, we extend the deterministic policy gradient (DPG) algorithm \cite{silver2014deterministic,lillicrap2015continuous,lowe2017multi} to our solution with improvements including 1) a clustering method to model a large number of merchants as multiple agents and 2) distributed architecture design to enable our framework to process requests in distributed workers in parallel and asynchronously.

\section{Taobao Display Ad System}

\begin{figure}[t]
\includegraphics[width=3in]{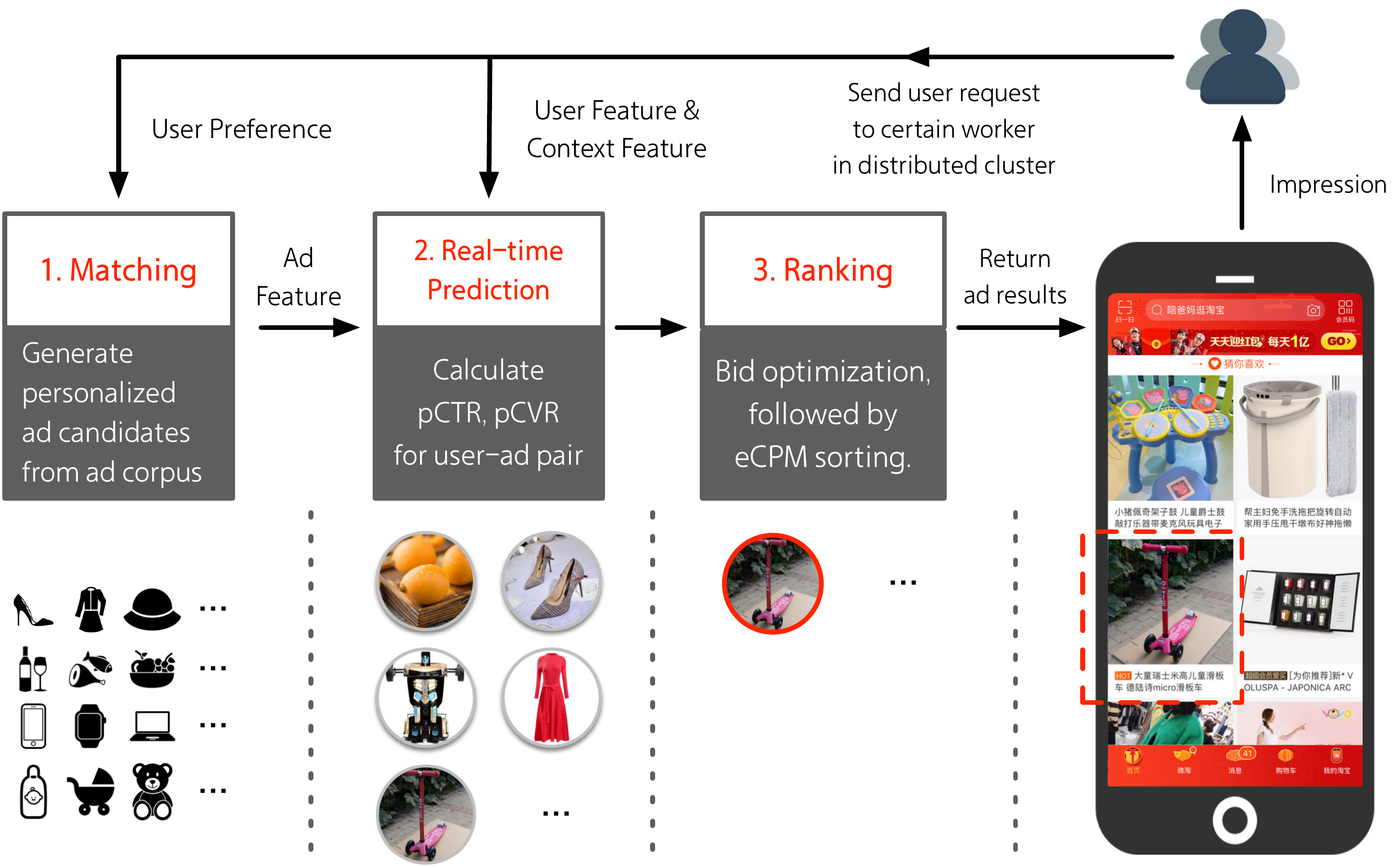}
\setlength{\abovecaptionskip}{0pt}
\caption{An Overview of Taobao Display Advertising System. Matching, RTP and Ranking modules sequentially process user requests, and finally return specified quantity of ads. These ads are shown in \emph{Guess What You Like} of Taobao App, tagged by \emph{Hot} (as shown in red dashed box) and surrounded with recommendation results.}
\label{fig:taobao_system}
\end{figure}

Taobao's advertisers are mostly the merchants who not only advertise but also sell products. Hereinafter, we call them merchants. Taobao ad system can be divided into three parts as shown in Figure \ref{fig:taobao_system}: First in the matching stage, user preferences are obtained by mining behavior data, and when receiving a user request, matching part recalls candidate ads (typically hundreds of ads) from the entire ad corpus in real time based on their relevancy. Different from recommender systems, the recall of the ads has to reflect the advertisers' willingness of bidding, i.e., their behavior targeting settings. Second, the follow-up real-time prediction (RTP) engine predicts the click-through rate (pCTR) and conversion rate (pCVR) for each eligible ad. Third, after real-time bidding for each candidate ad is received, these candidate ads are ranked by descending order of
$bid \times pCTR$, which is called effective cost-per-mille (eCPM) sorting mechanism. Finally, the ranked ads are displayed. For general RTB auction settings, we refer to \cite{wang2017display}.

The change of bids will influence the ranking of candidate ads, and further have the impact on the connections built between the consumers and merchants. An ideal mapping is that the consumers find their ideal products and the merchants target the right consumers who have the intent to buy the advertised products. When demands are precisely met by the supplies, the platform creates higher connection value for the society. For better revenue optimization, merchants authorize the platform to adjust their manually set bids within an acceptable range. In summary, bids as key control variables in the online advertising system and, if adjusted well, can achieve a win-win-win situation for all consumers, merchants and the platform's interest.

In our e-commerce system, there are a large number of registered merchants and registered consumers. Each auction is launched by a consumer. According to information in this auction, each merchant under its budget constraint gives a bid price. If a merchant wins an auction, the corresponding ad would be delivered to a consumer. This consumer has a probability to click the ad (click-through rate, CTR) to enter a landing page for the product, and then has a probability (conversion rate, CVR) to buy the merchant's product with price $ppb$ (pay-per-buy) forming the merchants' revenue. Given predefined budget to achieve higher revenue is a general goal of merchants. With the same predefined budget spent out, higher merchants' revenue means higher ROI ($ROI=revenue/budget$). Higher total merchants' revenue is also consumers' and platform's motivations: for consumers, they are connected to the products they want which means better consumer experience, while for the platform, larger gross merchandise volume (GMV) means larger long-term advertising revenue. Whenever a merchant's ad is clicked, the corresponding merchant's unspent budget is subtracted by advertising $cost$ according to generalized second price (GSP) auction with CPC mechanism \cite{edelman2007internet}. If a merchant loses an auction, he gets no reward and pays nothing. If the budget runs out, the merchant will not participate in any rest auctions.

Bidding in display advertising is often regarded as an episodic process \cite{cai2017real}. Each episode includes many auctions and each auction is about one consumer's page view in a very specific context. Auctions are sequentially sent to the bidding agents. Each merchant's goal is to allocate its budget for the right consumers at the right time to maximize its KPI such as revenue and ROI.
All the merchants competing together forms a multi-agent game. However, when budgets are limited, the game of merchants' bidding may result in a suboptimal equilibrium. For example, the merchants compete severely in early auctions and many merchants have to quit early, and the low competition depth in the late bidding results in low matching efficiency of consumers and merchants. Therefore, all merchants setting bids for different consumers in appropriate time according to various competition environments is essential for Taobao ad system to achieve a socially optimal situation.

\section{Multi-Agent Advertising Bidding}
\label{section:Approach}

We first formulate RTB as a Stochastic Game and then present our MARL approach and finally discuss our implementation details.

\subsection{RTB as a Stochastic Game}\label{section:RTB-MAMDP}
We formulate RTB as a \emph{Stochastic Game}, a.k.a. Markov Game \cite{fink1964equilibrium}, where there are $N$ bidding agents on behalf of merchants to bid ad impressions. A Markov game is defined by a set of states $\mathcal{S}$ describing the possible status of all bidding agents, a set of actions $\mathcal{A}_1,..., \mathcal{A}_N$ where $\mathcal{A}_i$ represents action spaces of agent $i$. An action $a \in \mathcal{A}_i$ is the bid adjustment ratio. According to $t$-th timestep state $s_t$, each bidding agent $i$ uses a policy $\pi_i : \mathcal{S}_i \mapsto \mathcal{A}_i$ to determine an action $a_i$ where $\mathcal{S}_i$ is state space of agent i. After the execution of $a_i$, the bidding agent $i$ transfers to a next state according to the state transition function $\mathcal{T}:\mathcal{S}\times \mathcal{A}_1\times...\times \mathcal{A}_N \mapsto \Omega (\mathcal{S})$ where $\Omega (\mathcal{S})$ indicates the collection of probability distributions over the state space. Each agent $i$ obtains a reward (i.e., revenue) based on a function of the state and all agents' actions as $r_i:\mathcal{S}\times \mathcal{A}_1\times...\times \mathcal{A}_N \mapsto \mathcal{R}$. The initial states are determined by a predefined distribution. Each agent $i$ aims to maximize its own total expected return $R_i=\Sigma_{t=0}^T \gamma^t r_i^t$ where $\gamma$ is a discount factor and $T$ is the time horizon. We describe the details of agents, states, actions, rewards and objective functions in our setting as follows.

\begin{figure}[t]
\includegraphics[width=3in]{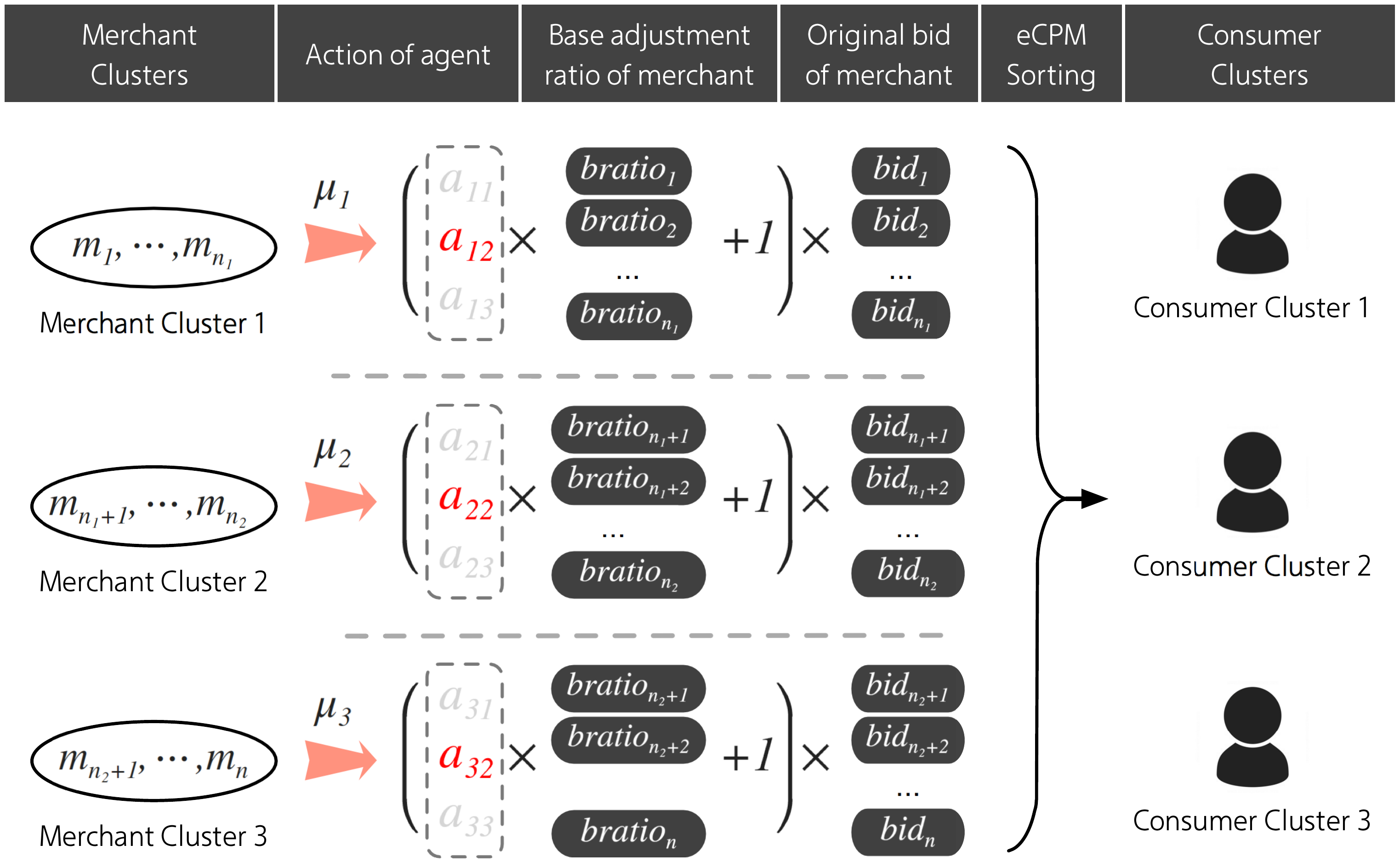}
\setlength{\abovecaptionskip}{5pt}
\caption{Merchants and consumers are grouped into clusters separately. Each merchant cluster is an agent, which adjusts ad bids of included merchants for different consumer clusters. For action $a_{ij}$, $i$ iterates the number of merchant clusters, as $j$ does for consumer clusters. $bratio_k$ stands for base adjustment ratio of merchant $k$. }
\label{fig:overview}
\end{figure}

\minisection{Agent Clusters}
In our system, $n$ registered merchants are denoted as $m_1, m_2,..., m_n$ and $l$ registered consumers are denoted as $c_1, c_2,..., c_l$. Each auction is launched by a consumer with a feature $x$ describing the consumer's information in this auction. The merchant's product's price is denoted as $ppb$ (pay-per-buy). The ideal way to formulate all merchants is to model each of them as an agent. However, such arrangement is computationally expensive, and in fact interactions between a specific consumer-merchant pair are very sparse. As the number of agents increases, the exploration noise becomes difficult to control. Thus, we propose a \emph{clustering method} to model the involved entities. With total revenue during one day as clustering feature, $n$ merchants are categorized as $N$ clusters $M_1,\ldots,M_N$. Similarly, with contributed revenue in one day as feature, $l$ consumers are categorized as $L$ clusters $C_1,\ldots,C_L$. We cluster consumers for building agents' states and for computing static features which enable the agents to evaluate features of auctions from different consumer clusters and adjust bids accordingly. Hereinafter, we use $i$ as subscript of merchant cluster, and $j$ for consumer cluster. Normally $N \ll n$, $L \ll l$, and when we shrink the cluster size and enlarge the cluster number, it approximates the ideal case. The diagram of this modeling is as Figure \ref{fig:overview}.

\begin{figure*}[t]
\includegraphics[width=6.5in]{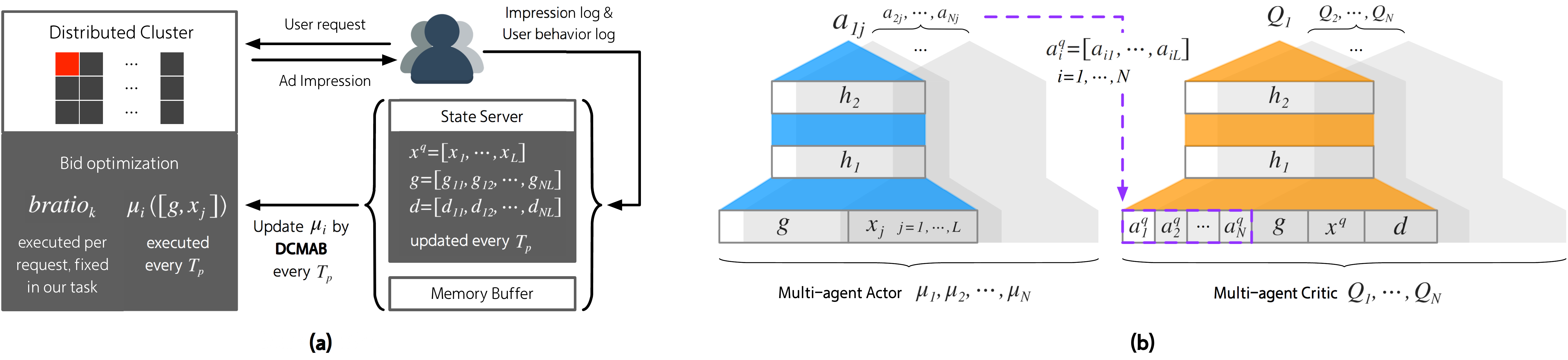}
\caption{DCMAB Illustration. \emph{(a)} DCMAB workflow in advertising system. The State Server maintains agents' states including general information $g$, consumer distribution $d$ and consumer static feature $x^q$. Every $T_p$, states are merged and agents' actors are updated. Then ${ \mu  }_{ i }\left( \left[ g,\, { x }_{ j } \right]  \right)$ is calculated for merchant cluster $i$ and consumer cluster $j$, further multiplied by $bratio_k$ to form final bid adjustment. \emph{(b)} DCMAB network. Separate Actor and Q network for each agent. ${ a }_{ ij }$ is calculated through ${ \mu  }_{ i }$ using $g$ and ${ x }_{ j }$ as input. In addition to states and actions, consumer distribution $d$ is collected as input of all agents' Q function. }
\label{fig:work_flow}
\end{figure*}

\minisection{State}
Our state design aims to let bidding agents optimize their budgets allocation based on both each impression's value and spending trends along time. We consider cumulative cost and revenue between merchants $M_i$ and consumers $C_j$ from the beginning of an episode up to now denoted as $g_{ij}\!=\!(cost_{ij},\!revenue_{ij})$ as the general information state. This is because all these $g_{ij}$ vectors characterize important information as:
(1) the budget spent status for an agent to plan for the rest auctions;
(2) the (cost, revenue) distribution of consumers for an agent to distinguish quality from different consumer clusters;
(3) the (cost, revenue) distribution of other agents for an agent to evaluate the competitive or cooperative environment.
Besides, the consumer feature $x$ is also added to the state which includes slowly-changed consumer features such as their total (cost, revenue) status updated every a period of time. This feature $x$ helps agents evaluate the auction better. We concatenate all $g_{ij}$ as $g = [g_{11},g_{12},\ldots,g_{NL}]$ with $x$ to form the state $s = [g,x]$. We suppose each merchant's budget is predefined, therefore their spent and unspent budgets information is maintained in the state. The diagram of this modeling is showed in Figure \ref{fig:work_flow}.

\minisection{Action}\label{section:action}
Every merchant manually sets different fixed bids for different consumer crowds. W.l.o.g., we denote the fixed bid as $bid_k$ across all the auctions, where $k$ iterates over $n$ merchants hereinafter. For better budget allocation, the platform is authorized to adjust $bid_k$ with a scalar $\alpha$ to generate final $\hat{bid}_k$ for execution, $\hat{bid}_k=bid_k\times(1+\alpha)$ where $\alpha\in [-range, range], 0<range<1$ and we use $range=0.9$ in our experiment. As stated above, we cluster $n$ merchants into $N$ clusters, then $\alpha$ should have $N$ different values for different merchant clusters. The actual bid adjust ratio used is $\alpha = { a }_{ i }\times { bratio }_{ k }$ as in Figure \ref{fig:overview}, where $a_i$ is the action of agent $i$ computed using learned neural networks and $bratio_k$ is impression-level feature to measure value of a specific impression for merchant $k$ such as pCVR calculated according to impression-level consumer-merchant information. The calculation of ${ bratio }_{ k }$ is predefined and we would discuss it in detail in \ref{seq:IDA}.

\minisection{Reward and Transition}
Reward is defined on the agent level. Cooperative and competitive relationships can be modeled with reward settings, i.e. competitive when every agent's reward is self-interested and cooperative when all agents' reward is the same. Taking competitive case as an example, when a merchant $k$ belonging to agent $i$ executes $\hat{bid}_k$ and wins an auction with delivering an ad to consumer of $C_j$, the reward of agent $i$ increases by the revenue (based on $ppb$) directly caused by this ad from this consumer. And after the ad was clicked, the budget of merchant $k$ decreases by $cost=pCTR_{next(k)}\times bid_{next(k)}/pCTR_k$ according to GSP mechanism where merchant $next(k)$ is the next ranked merchant of merchant $k$ according to maximum eCPM ranking score of $pCTR\times bid$. The $g_{ij}$ in state is updated by accumulating this $(revenue,cost)$. Changes of $g_{ij}$ for all $i,j$ including consumer feature $x$ changing form the transition of the states. If a merchant loses the auction, it contributes nothing to its agent's reward and state. Actually, our framework is able to use general reward such as revenue, cost, ROI, click, etc. In this paper, w.l.o.g., we consider revenue as our reward under fixed budget constraint, and we assume the merchant will spend out all his budget and use strategic bidding method to maximize his revenue. As $ROI=revenue/cost$ and cost is equal to this fixed budget, maximizing revenue also means maximizing ROI. Note that it's possible to maximize ROI by only choosing high ROI impressions and not bidding for low ROI impressions even there is money left in the budget, in which case although the merchant achieves a higher ROI, the revenue may be small, and this case is not considered in this paper.

\subsection{Bidding by Multi-Agent RL}
\label{sec:bbmarl}
Since the output action (bid adjustment) is in a continuous space, we adopt deterministic policy gradient for learning the bidding strategy. In the MARL setting, the Q function for agent $i$ is given as
\begin{align} 
Q_i^{\boldsymbol{\pi}}(s,\boldsymbol{a})=\mathbb{E}_{\boldsymbol{\pi},\mathcal{T}}[\Sigma_{t=0}^{T}\gamma^{t}r_i^t|s_0=s,\boldsymbol{a}],\label{eq:multiagentq}
\end{align}
where $\boldsymbol{\pi}=\{\pi_1,...,\pi_N\}$ is joint policy across all agents and $\boldsymbol{a}=[a_1,...,a_N]$ is joint action. $s_0$ is initial state. RL makes use of temporal difference recursive relationship with next time-step state $s'$ and joint action $\boldsymbol{a'}$ known as the Bellman equation:
\begin{align} 
Q_i^{\boldsymbol{\pi}}(s,\boldsymbol{a}) = \mathbb{E}_{r,s'}[r(s,\boldsymbol{a})+\gamma \mathbb{E}_{\boldsymbol{a'} \sim \boldsymbol{\pi}} [Q_i^{\boldsymbol{\pi}}(s',\boldsymbol{a'}]]. \label{eq:recurq-function}
\end{align}
When policy is deterministic, with a deterministic mapping function $\mu_i()$ from state $s$ to bidding action $a_i$ as Eq.(\ref{eq:dpmu}) for agent $i$ with parameter $\theta_i^{\mu}$. And $\mu_i()$ is commonly called \emph{actor}.
\begin{align}
a_i=\mu_{i}(s)=\mu_i([g,x])\label{eq:dpmu}
\end{align}
With Eq.(\ref{eq:dpmu}), above Eq.(\ref{eq:recurq-function}) becomes:
\begin{align} 
&Q_i^{\boldsymbol{\mu}}(s,a_1,...,a_N) \nonumber\\
&\quad\quad = \mathbb{E}_{r,s'}[r(s,a_1,...,a_N)+\gamma Q_i^{\boldsymbol{\mu}}(s',\mu_1(s'),...,\mu_N(s'))], \label{eq:q-function}
\end{align}
Where $\boldsymbol{\mu}=\{\mu_1,...,\mu_N\}$ is joint deterministic policy of all agents. In MARL, the goal is to learn an optimal strategy for each agent, which may have a different or even conflicted goal.  The notion of Nash equilibrium \cite{hu2003nashq} is important, which is represented as a set of policies $\mu^{*}=\{\mu_1^{*},...,\mu_2^{*}\}$ such that $\forall \mu_i$, it satisfies:

\begin{align} 
&Q_i^{\boldsymbol{\mu^{*}}}(s,\mu_1^{*}(s),...,\mu_N^{*}(s)) \nonumber\\
&\quad\quad = Q_i^{\boldsymbol{\mu^{*}}}(s,\mu_{i}^{*}(s),\mu_{-i}^{*}(s)) \ge Q_i^{\boldsymbol{\mu^{*}}}(s,\mu_{i}(s),\mu_{-i}^{*}(s)),\label{eq:nashq}
\end{align}

where we use compact notations for joint policy of all agents except $i$ as $\mu_{-i}^{*}(s)=\{\mu_1^{*}(s),...,\mu_{i-1}^{*}(s),\mu_{i+1}^{*}(s),...,\mu_N^{*}(s)\}$. In a Nash equilibrium, each agent acts with best response $\mu_i^{*}$ to others, provided all others follow policy $\mu_{-i}^{*}(s)$. This gives the optimal action at each state $s$ for agent $i$ and leads to equilibrium bidding strategy.

We solve $Q_i^{\boldsymbol{\mu^{*}}}$ and $\mu_{i}^{*}(s)$ in Eq.~(\ref{eq:nashq}) by using an alternative gradient descent approach, similar to the ones introduced in \cite{singh2000nash,lowe2017multi}, where we gradient update agent's $Q_i^{\mu}$ and $\mu_{i}(s)$ while fixing all other agent's parameters (thus their outputs). Specifically, the \emph{critic} $Q_i^{\mu}$ with parameter $\theta_i^{Q}$ is learned
by minimizing loss $L(\theta_i^Q)$  defined as
\begin{align}
L(\theta_i^Q) &= \mathbb{E}_{s,\boldsymbol{a},r,s'}[(Q_i^{\boldsymbol{\mu}}(s,a_1,...,a_N)-y)^2], \label{eq:q2} \\
y&=r_i+\gamma Q_i^{\boldsymbol{\mu'}}(s',\mu_1 '(s'),...,\mu_N '(s')), \label{eq:q1}
\end{align}
where $\boldsymbol{\mu'} =\{\mu_1 ',...,\mu_N '\}$ is target policies with delayed parameters $\theta_i^{\mu'}$, $Q_i^{\boldsymbol{\mu'}}$ is target critic function with delayed parameters $\theta_i^{Q'}$, and $(s,a_1,\ldots,a_N,r_i,s ')$ is a transition tuple saved in replay memory $D$. Each agent's policy $\mu_i$ with parameters $\theta_i^{\mu}$ is learned as
\begin{align}
\nabla_{\theta_i^{\mu}}J(\mu_i)=\mathbb{E}_{s} \big[\nabla_{\theta_i^{\mu}}\mu_i(s)\nabla_{a_i}Q_i^{\boldsymbol{\mu}}(s,a_1,...,a_N)|_{a_i=\mu_i(s)} \big]. \label{eq:learn_mu}
\end{align}

In the next section, we present a distributed implementation of Eqs.~(\ref{eq:q2}), (\ref{eq:q1}), and (\ref{eq:learn_mu}) within our distributed architecture.

\subsection{Implementation \& Distributed Architecture}
\label{seq:IDA}
A typical RL method such as original DPG saves a transition tuple after every state transition, which is difficult to implement in a real-world RTB platform for following reasons. (i) An operational RTB system consists of many distributed workers processing consumers' requests in parallel and asynchronously, demanding to merge all workers' state transitions. (ii) The states change frequently and saving every request as a transition tuple would cost unnecessary computation. In this section, we extend original gradient updates to be adapted to real-world distributed-worker platform.

The state transition update and action execution are maintained asynchronously. In other words, transition tuples and executing actions are operated with different frequencies, where states that merge among workers and tuples are saved periodically every a time gap $T_p$. During each $T_p$, there are many requests processed. For every request, according to different request features, the actor $\mu_i$ generates different actions for execution. With our method, the merge of states and transition updates at every $T_p$ interval can be handled by current industrial computation ability of distributed workers. Note that although states are updated every $T_p$, the actions are generated for every auction in real time. This framework brings different frequencies of critic updates and actor executions. We propose following techniques to organize critic and actor well.

\minisection{Balance Computing Efficiency and Bid Granularity} For computing efficiency, states are updated every $T_p$. For finer bid granularity, we introduce impression-level feature ${ bratio }_{ k }$ to fine tune bid price. As stated in State definition, consumer feature $x$ consists of static feature containing slowly changed information obtained before one $T_p$ starts. And real-time feature such as $pCVR$ is also utilized, which can only be acquired when a request visits the worker.

As shown in actor definition, we factorize the final bid adjustment as $\alpha={ a }_{ i }\times { bratio }_{ k }$. ${ a }_{ i }$ is computed every $T_p$ by ${ \mu  }_{ i }\left( \left[ g,\, x \right]  \right)$, where $x$ is static consumer feature. While the real-time part is used for ${ bratio }_{ k }$ in every impression. The concrete formulation is ${ bratio }_{ k }=pCVR_k/pCVR_k^{avg}$, where $pCVR_k$ is on merchant-consumer level (not merchant cluster and consumer cluster level) for merchant k and $pCVR_k^{avg}$ is 7-day historical average $pCVR_k$ of this merchant k. $pCVR_k/pCVR_k^{avg}$ provides impression-level information enabling a merchant to apply impression-level bid adjustment for high quality consumer request as Zhu et al. \cite{zhu2017optimized}. In such settings, ${ a }_{ i }$ applies coarse adjustment to merchants within a cluster, and ${ bratio }_{ k }$ discriminates among merchants within the same cluster and reflects real-time conversion value.

Next, we focus on the learnable component $a_i$, i.e. $\mu_i$. Computing $\mu_i([g,x])$ for every consumer is computationally costly before every time interval $T_p$ because of large numbers of consumers. Our solution is utilizing consumer clusters. For $L$ consumer clusters, we design $L$ cluster-specific versions of features for $x$ as $x^q=[x_{1},...,x_{L}]$. Each $x_{j}$ contains a one-hot embedding of consumer cluster $j$ with dimension $L$, and its historical $(revenue, cost)$. We design this one-hot embedding to enhance the discriminative ability on the basis of $(revenue, cost)$. Before the beginning of each $T_p$, we compute $a_{ij}=\mu_{i}([g,x_{j}])$ for every merchant cluster $i$ and consumer cluster $j$ pair for $i=1,...,N, j=1,...,L$. Within one interval $T_p$, for candidate ad of merchant k, we select $a_{ij}$ according to the merchant cluster and consumer cluster pair, then multiplied by ${ bratio }_{ k }$ and clipped by $[-range, range]$ to form final adjusting ratio $\alpha=\min\{ \max\{a_{ij}\times { bratio }_{ k },-range\}, range\}$ for computing $\hat{bid}_k = bid_k \times (1+\alpha)$. Note that $a_i$ and $x$ in Eq.(\ref{eq:dpmu}) are replaced by $a_{ij}$ and $x_j$ due to extra dimension of consumer cluster.

\minisection{Handle Impression-Level Information Summarization}
We save transition tuples to replay memory every time interval $T_p$, which requires to aggregate all impressions' information during $T_p$. Thus, we propose an aggregation method to summarize the executions within $T_p$ where we maintain a discrete distribution of $a_{ij}$ as $d_{ij}={\# a_{ij}}/{tot\_num}$ where ${\# a_{ij}}$ stands for executed number of $a_{ij}$ and ${tot\_num}$ for all executed number. We concatenate all $d_{ij}$ as $d=[d_{11},d_{12},...,d_{NL}]$ and save $d$ as a part of tuple every $T_p$. And the critic function $Q$'s input is augmented as $Q(s^q,a_{1}^q,...,a_{N}^q,d)$ where $s^q=[g,x^q]$ and $a_i^q=[a_{i1},...,a_{iL}]$.

Our distributed gradient update aims to let agents optimize budgets allocation according to consumer distributions and consumer features every $T_p$ while utilizing real-time feature such as $pCVR$ in every impression. We call our algorithm Distributed Coordinated Multi-Agent Bidding (DCMAB) with critic and actor update rules:
\begin{align}
&y =r_i+\gamma Q'_i({s^q}{'}, {a_1^q}{'},...,{a_N^q}{'}, d') \big|_{{a_o^q}{'}=[\mu_{o}'([g',x_1']),...,\mu_{o}'([g',x_L'])]}\label{eq:oury}\\
&L(\theta_i^Q) =(y-\gamma Q_i(s^q, a_1^{q},...,a_N^{q},d))^2\label{eq:ourQ}\\
&\nabla_{\theta_i^{\mu}}J\approx \Sigma_j \nabla_{\theta_i^{\mu}}\mu_i([g,x_j]) \nabla_{a_{ij}^{q}}Q_i(s^q, a_1^{q},...,a_N^{q},d)\label{eq:ourmu}
\end{align}

The solution is as Figure \ref{fig:work_flow} and pseudo code as Algorithm \ref{alg:DCMAB}.

\minisection{Online-Like Offline Simulator} An offline simulator can significantly accelerate reinforcement learning algorithm research. Considering follow-up online deployment of the algorithm, we developed an offline simulator whose upstream and downstream data flow environments are identical to online engine, and its distributed-worker design can meet the online service engineering requirement. All experiments in this paper is based on this offline simulator. Our current move is transferring the offline system to online and deploying our DCMAB algorithm for online A/B testing.

\begin{algorithm}[t]
    \caption{DCMAB Algorithm}\label{alg:DCMAB}
    Initialize $Q_i(s^q, a_1^q,...,a_N^q,d|\theta_i^Q)$, actor $\mu_i([g,x]|\theta_i^{\mu})$, target network $Q_i '$, $\mu_i '$ with $\theta_i^{Q'}\leftarrow \theta_i^Q$, $\theta_i^{\mu'}\leftarrow \theta_i^{\mu}$ for each agent $i$.\\
    Initialize replay memory $D$\\
    \For{episode = 1 to $E$}{
        Initialize a random process $\mathcal{N}$ for action exploration\\
        Receive initial state $s$ for all agents\\
        \For{$t$ = 1 to $T$}{
            For each agent $i$, compute $a_i^q$ and add $\mathcal{N}_t$.\\
            \For{auctions in parallel workers in $T_p$}{
                For each agent $i$, compute $bratio$ and combined with $a_i^q$ compute adjusting ratio $\alpha$ and execute.\\
                For each agent $i$, save reward, cost and maintain distribution $d$.
            }
            For each agent $i$, merge rewards, cost in last $T_p$ to get reward $r_i$ and update state to $s^q{'}$. Store $(s^q,d,a_i^q,r_i,s^q{'})$ to replay memory.\\
            $s^q{'} \leftarrow s^q$

            \For{agent $i$=1 to N}{
                Sample a random minibatch of $S$ samples $(s^q,d,a_1^{q},...,a_N^{q},r_i,s^q{'},d')$ from $D$\\
                Update critic by minimizing loss with Eqs.(\ref{eq:oury}),(\ref{eq:ourQ}).\\
                Update actor with Eq.~(\ref{eq:ourmu}).\\
                Update target network: $\theta'\leftarrow\tau\theta+(1-\tau)\theta$\\
            }
        }
    }
\end{algorithm}

\section{Experiments}\label{section:ExperimentalSetup}

Our experiments are conducted over the data sets collected from Taobao display ad system. The data is collected in \emph{Guess What You Like} column of Taobao App Homepage where three display ads slots and hundreds of recommendation slots are allocated. As we have collected the bid prices traded in the market as well as the feedback and conversions from the consumers for the placed ads, we would be able to replay the data to train and test our proposed DCMAB in an offline fashion. The similar settings can be found in other research work \cite{cai2017real,wang2017ladder,zhu2017optimized,perlich2012bid,zhang2014optimal}.

\subsection{Data Sets and Evaluation Setup}

\minisection{Data sets}
The data sets we used for experiments come from real-world production. The display ads are located in \emph{Guess What You Like} column of Taobao App Homepage where three display ads slots and hundreds of recommendation slots are well organized. Based on the log data, the saved procedures of consumers' requests including pCTR, pCVR, ppb along with the requests are used as procedure replay to form an offline simulation platform. And pCTR, pCVR, ppb are used to simulate the consumers' behaviors for computing states and rewards. We use the $1/20$ uniformly sampled first three hours' logged data from date of 20180110 as training data, and the $1/20$ uniformly sampled first three hours' logged data from 20180111 as test data. Training and test of our algorithm are both based on the offline simulation system due to the lack of real consumer feedback data. All results reported are based on test data.

For merchants, when budget is unlimited, each merchant will adjust bid price to the highest number and the solution is trivial. To test optimized budget allocation along time, the budget for each merchant should not be too large. Similar to the setting in \cite{zhang2014optimal}, we determine the budget as follows: let all merchants use manually set bid with unlimited budgets and accumulate the total cost $C_T$. Then each merchant's budget is set as a fraction of $C_T$.

With notion $C_T$, here are some statistics of the data: for training set there are 203,195 impressions, 18,532 revenue ($C_T$) and 5,300 revenue ($C_T/3$) where ($C_T$) means the setting where merchants are endowed with unlimited budgets (in real situation this is impossible, when merchants have limited budgets, they quit bidding when budgets run out and the market depth decreases); for testing set there are 212,910 impressions, 18,984 revenue ($C_T$) and 5,347 revenue ($C_T/3$); for both data sets, there are 150,134 registered consumers and 294,768 registered merchants. All revenue unit is CNY.

\minisection{Evaluation metrics}
Evaluation is based on agents' revenue, ROI and CPA (cost per acquisition), and total traffic revenue, ROI and CPA under predefined budgets and a number of auctions. We define CPA as $CPA=cost/click$. The agent's objective is to maximize its revenue given the budget. We also analyze the influences of the agents' rewards changes on the converged equilibrium.

\minisection{Evaluation flow}
We built an offline simulator close to the real online system with distributed workers processing consumers' requests. As stated in Section \ref{section:action}, with $range=0.9$, the feasible bid region is $bid*(1+\alpha)\in[0.1*bid, 1.9*bid]$ where bid is a merchant's original bid and $\alpha$ is optimized by solving Eq.~(\ref{eq:nashq}) using Eq.~(\ref{eq:oury})(\ref{eq:ourQ})(\ref{eq:ourmu}) as in Algorithm~\ref{alg:DCMAB}. In each auction, according to the maximum eCPM ranking, the top-ranked three merchants win. During our training, as model learns, the model's different bids lead to different ranking results. Due to lack of consumers' real feedback of all different ranking results for all merchants, we use expected CPC ($cost_k\times pCTR_k$ where $cost_k=pCTR_{next(k)}\times bid_{next(k)}/pCTR_k$ is based on GSP mechanism) and expected revenue ($pCTR_k\times pCVR_k \times ppb_k$) for offline simulation. The system is based on 40-node cluster each node of which has Intel(R) Xeon(R) CPU E5-2682 v4, 2.50GHz and 16 CPU cores with 250 GB memory on CentOS. The model is implemented with distributed TensorFlow. Our offline platform is consistent with the online platform, in online deployment we only need to change the reward from expectation to real feedback.

\minisection{Episode length}
To simulate the real online system, our simulation platform updates states every hour. We use three hours' auctions for evaluation. The length of an episode includes three steps which is the number of state transitions. The three-hour training data includes 203,195 impressions which is the number of actor executions. Each training task takes about 4 hours with 40 distributed workers.

\subsection{Compared Methods}
With same settings, following algorithms are compared with our DCMAB. Except manually set bids, all other algorithms use neural networks as approximators. We also build a reward estimator for contextual bandit as a critic. All algorithms' critics include two hidden layers with 100 neurons for the first hidden layer and 100 neurons for the second hidden layer with states as inputs to the first layer and actions as inputs to the first hidden layer. All algorithms' actors include 300 neurons for the first hidden layer and 300 neurons for the second hidden layer with states as inputs and actions as outputs. The activation function for hidden layers is \emph{relu}, $tanh$ for output layer of actors and linear for output layer of critics.

\begin{itemize}[leftmargin=2.5mm]
  \item \textbf{Manually Set Bids.} They are the real bids set manually by human according to their experiences.
  \item \textbf{Contextual Bandit.} This algorithm \cite{lilchu2010contexbandit} optimizes each time step independently. Each impression's bid is adjusted according to only the feature in the impression (contextual feature). To compare with DCMAB, we also add other agents' actions as parts of contextual feature. The key difference between this algorithm and ours is that it doesn't optimize budgets allocation along time.
  \item \textbf{Advantageous Actor-critic (A2C)} This \cite{mnih2016asynchronous,sutton1998reinforcement,degris2012mfrlcap} is an on-policy actor-critic algorithm without a memory replay. The critic function $Q$ of A2C doesn't take other agents' actions as input.
  \item \textbf{DDPG.} DDPG \cite{lillicrap2015continuous} is an off-policy learning algorithm with a memory replay. The critic function $Q$ of this algorithm doesn't take other agents' actions as input.
  \item \textbf{DCMAB.} This is our algorithm. We upgrade MADDPG \cite{lowe2017multi} with clustered agents modeling and redesign actor and critic structures to adapt to distributed workers' platform. The critic function $Q$ of this algorithm takes all agents' actions as input.
\end{itemize}

\subsection{Hyperparameter Tuning}
\subsubsection{Clustering Method}\label{sec:clustermethod}
\begin{figure}[t]
\includegraphics[width=3.5in]{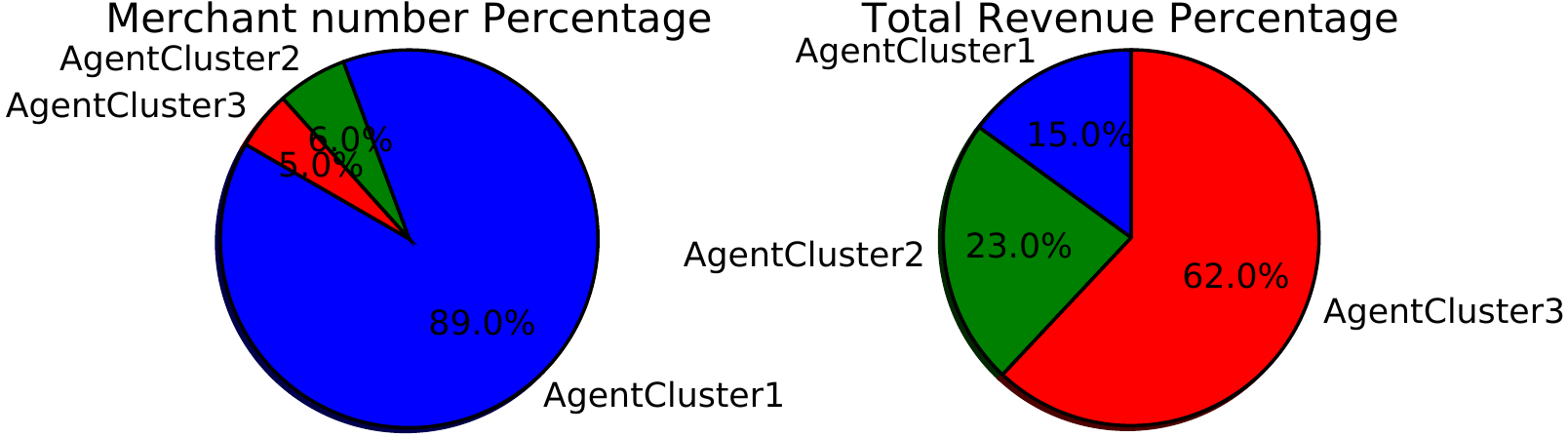}
\caption{Clusters of Merchants}
\label{fig:cluster_pie}
\end{figure}

When a consumer request comes, according to its requested marchant criteria, our system firstly selects $n_c$ candidates of merchants from all $n$ registered merchants where $n_c \ll n$. And these $n_c$ candidates attend the bidding stage while other $n-n_c$ merchants are filtered out. We consider one merchant who is present in bidding stage as one presence. We rank all $n$ merchants according to their revenues in training data and group them into clusters where these clusters have approximately equal presences respect to all consumer requests in training data. This clustering method makes the competitions among agent clusters relatively balanced. The example of three clusters is as Figure \ref{fig:cluster_pie}. Usually, clusters with higher revenues consist of small numbers of merchants and contribute larger amount of revenue. The reason is that most high-revenue merchants attend the bidding stage more frequently. Consumers are also ranked according to their revenues and grouped into clusters with each cluster having equal proportion of requests to the ad platform. Note that it's possible to cluster merchants and consumers with more features than only revenue according to specific business needs. This paper considers revenue as an example method. This clustering preprocessing is always done before training procedure according to recent log data to ensure the clustering principle is up-to-date.

\subsubsection{Number of Clusters}
In our formulation, theoretically, more clusters and smaller cluster sizes provide more possible adjusting ratios meaning better possible solutions. We tried different cluster numbers $\{1,2,3,4,5,10,30\}$ as Figure \ref{fig:budgetSearch}(a). Two kinds of rewards are used. 'Coord' means all clusters' rewards are the same as total traffic revenue. 'Self-Interest' means each cluster's reward is its own revenue. For both rewards, we use total traffic revenue as metric.

In Figure \ref{fig:budgetSearch}, horizontal axis is the number of agent clusters, and vertical axis represents total traffic revenues. We draw the mean episode reward as blue and red curves with corresponding colored area as standard deviations. From the results, we find the best performance is achieved when the number of clusters is 3 and 4. When cluster number increases from 1 to 3, the performance increases showing the benefits of shrinking cluster size and adding more clusters. When we further increase cluster number from 4 to 30, we find the performance drops. We observe as we increased the number of agents, the agents' policies learning easily converged to worse equilibria as many agents competed severely in early stage with high bid prices and quited auctions earlier. There exists better strategies for these agents such as lowering bids in early stage and competing for cheaper auctions in late stage. After cluster number tuning, cluster number 3 appears to perform the best, and our follow-up experiments shall fix the number of clusters as 3.

\subsubsection{Budget Search}
\begin{figure}[t]
\includegraphics[width=3.4in]{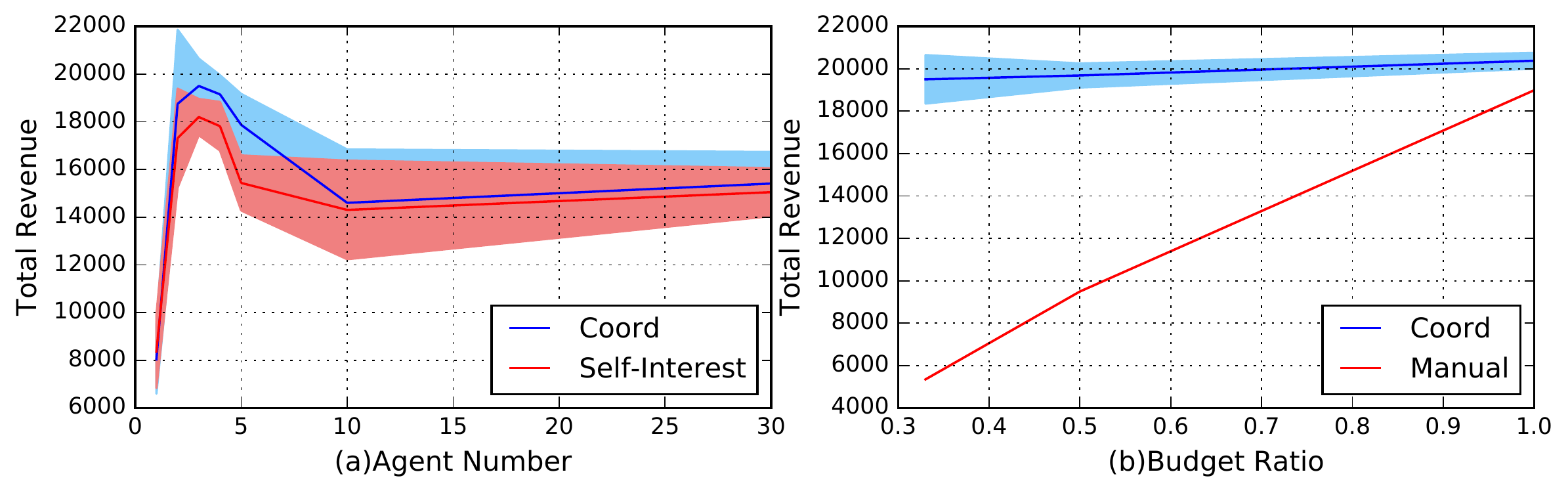}
\caption{(a) Revenue(CNY) vs. Agent Number \\(b) Revenue(CNY) vs. Budget Ratio}
\label{fig:budgetSearch}
\end{figure}

With the three agent clusters fixed, we now measure the total revenue performance of our DCMAB with manually set bids shown in Figure \ref{fig:budgetSearch}(b) where 'Coord' means all agents' rewards are total revenue. The budget for each merchant is searched from one-third to full amount of unlimited budget and in all cases over 99\% of the given budget is spent out, which means higher revenue is always better. Compared with manually setting, our DCMAB with coordinated rewards consistently maintain a higher revenue even when budget is low due to the better budget allocation. Manually setting bids acquires more revenue as the budget increases because higher budget makes more merchants stay in the market and deliver their ads to the consumers.

\subsection{Experimental Results}\label{section:ExperimentalResults}
In this section, we compare our DCMAB algorithm with the baselines to understand their learning abilities and performance.

\subsubsection{Performance Comparisons}

\begin{table*}[t]
  \caption{ROI/CPA/COST from Self-Interest Bidding Agents}
  \label{tab:baseline_roi}
  \resizebox{1.9\columnwidth}{!}{
    \begin{tabular}{c|ccc|ccc|ccc|ccc}

      \toprule
      &  \multicolumn{3}{|c|}{AgentC1}  &  \multicolumn{3}{|c|}{AgentC2}  &  \multicolumn{3}{|c|}{AgentC3}  &  \multicolumn{3}{|c}{Total} \\
      \midrule
      Indices  &  ROI&CPA&COST  &   ROI&CPA&COST  &  ROI&CPA&COST & ROI&CPA&COST\\
      \midrule
      Manual&100.00\%&100.00\%&99.65\% & 100.00\%&100.00\%&99.71\% & 100.00\%&100.00\%&99.42\% & 100.00\%&100.00\%&99.52\%\\
      Bandit&121.38\%&82.43\%&99.87\% & 159.41\%&62.73\%&99.53\% & 102.63\%&97.39\%&99.64\% & 112.14\%&84.23\%&99.63\%\\
      A2C &103.30\%&96.57\%&99.39\% & 106.85\%&93.58\%&99.60\% & 170.91\%&58.55\%&99.66\% & 158.38\%&68.09\%&99.62\%\\
      DDPG &577.87\%&17.27\%&99.51\% & 976.80\%&10.23\%&99.18\% & 164.29\%&60.85\%&99.76\% & 305.75\%&24.26\%&99.58\%\\
      DCMAB&690.18\%&14.46\%&99.38\% & 584.63\%&17.10\%&99.43\% & 275.11\%&36.34\%&99.57\% & 340.38\%&24.84\%&99.51\%\\
      \bottomrule
    \end{tabular}
  }
\end{table*}

\begin{figure*}[t]
\includegraphics[width=0.9\textwidth]{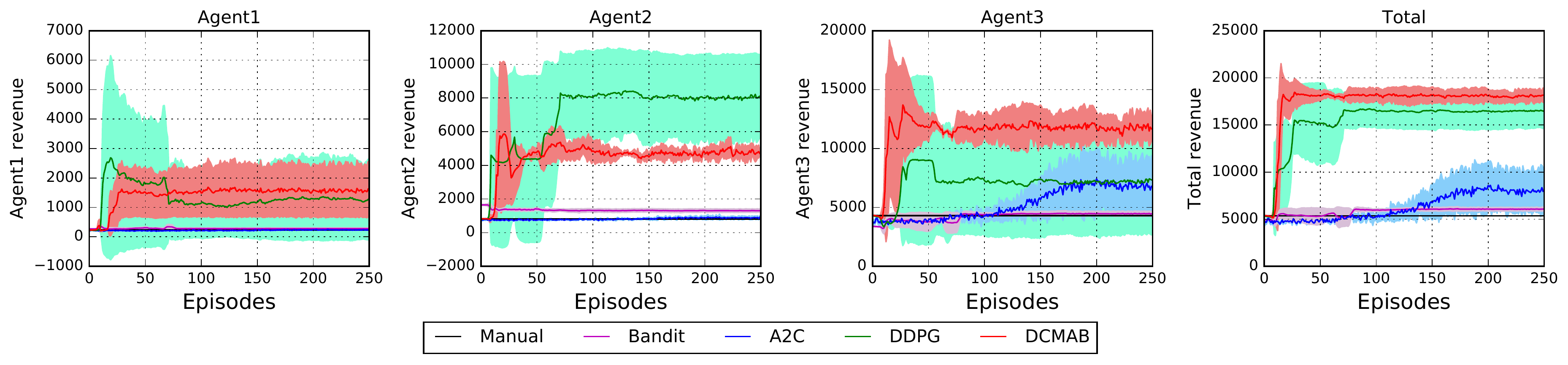}
\caption{Learning Curves Compared with Baselines. Revenue unit: CNY}
\label{fig:baseline}
\end{figure*}
For performance test, we set the best hyperparameters as tuned in the previous section. For instance, we group merchants and consumers into 3 clusters, respectively. Each merchant's budget is set as $C_T/3$. Each agent cluster's reward is set as its own episode revenue, which is a self-interest reward. The results are reported in Figure \ref{fig:baseline}, Table \ref{tab:baseline} and Table \ref{tab:baseline_roi}.

Table \ref{tab:baseline} lists the converged performances of different algorithms (we consider the training performance not improving in last 50 episodes as converged). Each row shows an algorithm's results. The columns represent the results of different agent clusters' and their summed total revenue in one algorithm's experiment. We conducted 4 times of experiments for each algorithm and gave the average revenues and standard deviations in Table \ref{tab:baseline}.

We use Pareto improvement \cite{fudenberg1991game} as one cluster can improve its revenue without hurting other clusters' revenues. Among all algorithms, our DCMAB has Pareto improvement over all other algorithms except DDPG, which means all clusters' revenue and total revenue are improved. This verifies the effectiveness of our algorithm. DDPG has Pareto improvement than Manual and Bandit. Compared with on-policy algorithm A2C, DDPG and our DCMAB perform better, illustrating the usefulness of sample memory. Compared with Bandit, other algorithms as A2C, DDPG and our DCMAB verify the importance of budget allocation among different hours, which points out the necessity of reinforcement learning modeling rather than bandit modeling. Manually setting bids perform the worst as it is a non-learning baseline.

DCMAB and DDPG result in different equilibria. AgentC1 and AgentC3 get more revenue in DCMAB than in DDPG while AgentC2 gets more revenue in DDPG than in DCMAB. Comparing these two equilibria, we find DCMAB achieves a higher total revenue of 18199 than DDPG of 16359. From perspective of total matching efficiency for connecting consumers to products, DCMAB gives better results. Moreover, DCMAB gives a more stable equilibrium with all agents' revenues and total revenue's standard deviation lower than DDPG, which verifies the merits of modeling all agents' actions in DCMAB rather than only modeling own action in DDPG.

Table \ref{tab:baseline_roi} lists ROI, CPA normalized respect to manual bids of all agents and their summation. ROI is defined as $ROI = revenue/cost$ where revenue is merchants' income and cost is the money paid to the platform by merchants. CPA is defined as $cost/click$ where click is the total click numbers from the consumers which is computed as $click = \sum pCTR$ in our offline simulation. Table \ref{tab:baseline_roi} COST columns show cost spent percentage ($cost/budget$), we find almost all agents' cost spent out which is reasonable for competing for more revenue under constrained budgets. ROI columns show DCMAB achieves highest ROI in AgentC1, AgentC3 and Total, and CPA columns present DCMAB costs less money for same numbers of click in AgentC1, AgentC3 and Total, which demonstrates ROI and CPA optimization ability of DCMAB.

The learning is illustrated in Figure \ref{fig:baseline}. We find our DCMAB converges more stable than DDPG, verifying the effectiveness of modeling all agents' actions as inputs to action-value functions. DCMAB and DDPG learn faster than A2C and bandit, showing the merits of the deterministic policy gradient with a memory replay.

\begin{table}[t]
	\caption{Revenue(CNY) from Self-Interest Bidding Agents}
	\label{tab:baseline}
	\resizebox{1\columnwidth}{!}{
		\begin{tabular}{c|ccc|c}

			\toprule
			&  AgentC1  &  AgentC2  &  AgentC3  &  Total \\
			\midrule
			Manual  &  231  &   817 & 4299  &  5347 \\
			Bandit  &  281$\pm$21  &  1300$\pm$50 & 4422$\pm$171  &  6003$\pm$123 \\
			A2C     &  238$\pm$7  &   872$\pm$104 & 7365$\pm$2387  &  8477$\pm$2427 \\
			DDPG    & 1333$\pm$1471  &  7938$\pm$2538 & 7087$\pm$4311  & 16359$\pm$1818 \\
			DCMAB  & 1590$\pm$891  &  4763$\pm$721 & 11845$\pm$1291 & 18199$\pm$757 \\
			\bottomrule
		\end{tabular}
	}
\end{table}

\subsubsection{Coordination vs. Self-interest}
\begin{figure}[t]
\includegraphics[width=3.3in]{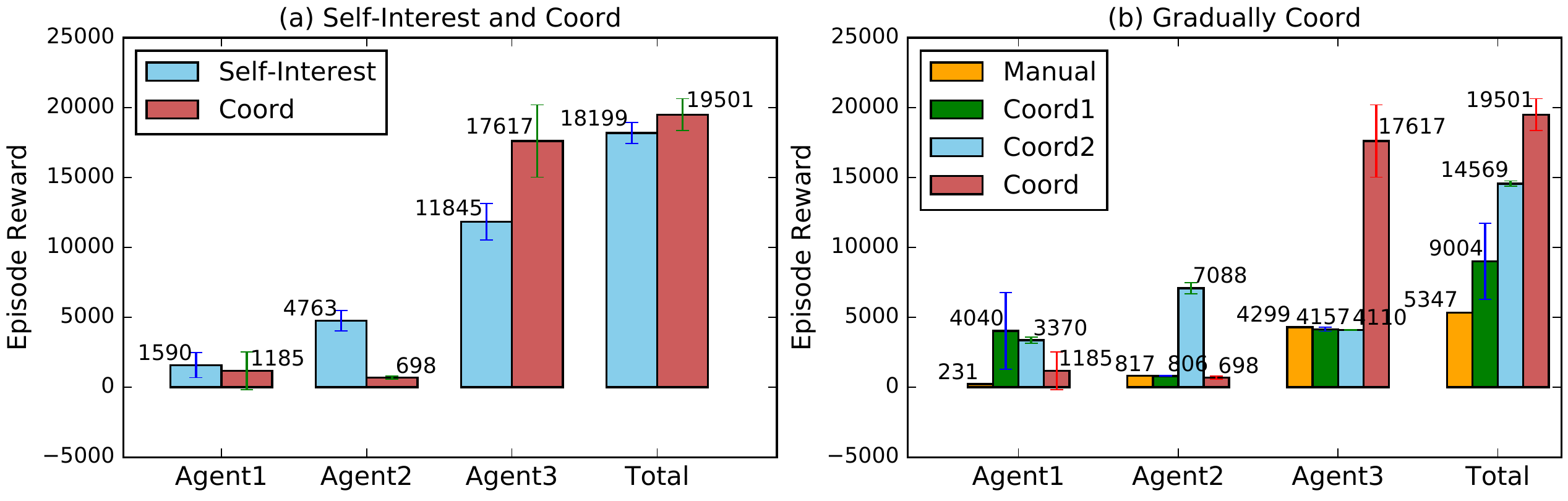}
\caption{(a): Self-Interest VS. Coord; (b): Gradually Coord\\ Episode Reward: revenue(CNY)}
\label{fig:coordination}
\end{figure}

This part studies how different reward settings influence the equilibrium reached when agents optimize revenue with all budgets spent out. First, we compare two kinds of reward settings as Table \ref{tab:coord_2} and Figure \ref{fig:coordination}(a). Self-Interest stands for each agent reward set with its own revenue; Coord stands for all agents' rewards set as total traffic revenue where all agents are fully coordinated to maximize the same goal. We find Coord achieves better total revenue than Self-Interest. Compared to the Self-Interest equilibrium, in Coord's equilibrium, while Agent1 and Agent2 obtain less revenues, Agent3's revenue is improved largely, resulting in a total revenue improvement. The total revenue improvement of Coord shows the ability of DCMAB to coordinate all agents to achieve a better result for overall social benefits.

In Table \ref{tab:coord} and Figure \ref{fig:coordination}(b), we analyze the performance when we gradually add learned agents' bids with coordination reward while keeping other agents' bids manually set. In Figure \ref{fig:coordination}(b), Manual means all agents are self-interested with manually set bids; Coord1 stands for that only bids of agent cluster 1 are learned with total revenue reward while other two agents' bids are manually set; Coord2 stands for Agent1 and Agent2's bids are learned with rewards of total revenue while Agent3's bids are manually set; Coord means all agents' bids are learned with rewards of the total revenue.

Compared to Manual,  the total revenue of Coord1 setting is improved from 5347 to 9004. The improvement mainly comes from Agent1 (from 231 revenue to 4040 revenue), while Agent2 (817 to 806) and Agent3 (4299 to 4157) do not contribute to the total improvement. This illustrates that the flexibility of the MARL framework from our approach in adjusting the coordination level depending on the specific needs in practice.

With Coord2, total revenue is improved more than Coord1 and it mainly comes from Agent1 (from 231 to 3370) and Agent2 (from 817 to 7088) while Agent3 drops a little. As more merchants join the cooperation, total revenue is further improved from Coord1 of 9004 to Coord2 of 14569. By comparing Coord2 and Coord1, we find Agent2's revenue increases largely from 806 to 7088, while Agent1's revenue unfortunately drops from 4040 to 3370. This shows Coord2 rearranges traffic allocation and would inevitably harm the performance of some agents to achieve better overall revenue.

Finally, when all agents cooperate for total revenue, it achieves the highest total revenue. As all agents' rewards aim at total revenue, we find Agent1 and Agent2 reach a compromise with dropped revenue compared to Coord1 and Coord2. And Coord rearranges the traffic to unleash Agent3's potential to improve the total revenue resulting in a larger improvement of total revenue from Coord2 14569 to 19501. In terms of total revenue, from Coord1, Coord2 to Coord, the gradually added coordination verifies our DCMAB's ability to reinforce all agents to cooperate for a predefined global goal. From a system perspective, higher total revenue means the consumers' better experiences for better connections to the commodities they like. From a long-term perspective, maximizing total revenue also encourages merchants to improve their business operational efficiency and provide better products to consumers.

\begin{table}[t]
  \caption{Revenue(CNY) of Self-Interest/Full Coordination}
  \label{tab:coord_2}
  \resizebox{1\columnwidth}{!}{
  \begin{tabular}{c|ccc|c}
    \toprule
                                &  Agent1   &  Agent2  &  Agent3   &  Total \\
    \midrule
    Self-Interest              &  1590$\pm$891  &  4763$\pm$721 & 11845$\pm$1291  & 18199$\pm$757 \\
    Coord                &  1185$\pm$1359  &   698$\pm$100 & 17617$\pm$2583  & 19501$\pm$1144 \\
  \bottomrule
\end{tabular}
}
\end{table}

\begin{table}[t]
  \caption{Revenue(CNY) for Different Coordination Levels}
  \label{tab:coord}
  \resizebox{1\columnwidth}{!}{
  \begin{tabular}{c|ccc|c}

    \toprule
                                &  Agent1   &  Agent2  &  Agent3   &  Total \\
    \midrule
    All Manual              &   231  &   817 &  4299  &  5347 \\
    1 PartiallyCoord        &  4040$\pm$2732  &   806$\pm$28 &  4157$\pm$145  &  9004$\pm$2728 \\
    2 PartiallyCoord  &  3370$\pm$218  &  7088$\pm$395 &  4110$\pm$16  & 14569$\pm$195 \\
    Fully Coord                &  1185$\pm$1359  &   698$\pm$100 & 17617$\pm$2583  & 19501$\pm$1144 \\
  \bottomrule
\end{tabular}
}
\end{table}

\section{Conclusions}\label{section:Conclusions}
In this paper, we proposed a Distributed Coordinated Multi-Agent Bidding solution (DCMAB) for real-time bidding based display advertising.
The MARL approach is novel and for the first time  takes into the interactions of all merchants bidding together to optimize their bidding strategies. It utilizes rich information from other agents' actions, the features of each historic auction and user feedback, and the budget constraints etc.
Our DCMAB is flexible as it can adjust the bidding that is fully self-interested or fully coordinated. The fully coordinated version is of great interest for the ad platform as a whole because it can coordinate the merchants to reach a better socially-optimal equilibrium for balancing the benefits of consumers, merchants and the platform all together.
We realized our model in a product scale distributed-worker system, and integrated it with the process auctions in parallel and asynchronously. Experimental results show that our DCMAB outperforms the state-of-the-art single agent reinforcement learning approaches. With fully cooperative rewards, DCMAB demonstrates its ability of coordinating all agents to achieve a global socially better objective.
As the results from the offline evaluation are promising, we are in process of deploying it online. We plan to conduct live A/B test in Taobao ad platform with a particular focus on mobile display ads.

\begin{acks}
The authors would like to thank Jian Xu, Qing Cui and Lvyin Niu for their valuable help and suggestions.
\end{acks}

\bibliographystyle{ACM-Reference-Format}
\bibliography{sample-bibliography}

\end{document}